\documentclass[10pt,twocolumn,letterpaper]{article}
\usepackage[T1]{fontenc}
\usepackage[utf8]{inputenc}
\usepackage{color}
\usepackage{array}
\usepackage{float}
\usepackage{multirow}
\usepackage{amsmath}
\usepackage{graphicx}
\usepackage{esint}
\usepackage[unicode=true,
 bookmarks=false,
 breaklinks=false,pdfborder={0 0 1},backref=page,colorlinks=true]
 {hyperref}

\makeatletter

\providecommand{\tabularnewline}{\\}
\floatstyle{ruled}
\newfloat{algorithm}{tbp}{loa}
\providecommand{\algorithmname}{Algorithm}
\floatname{algorithm}{\protect\algorithmname}

 \usepackage{cvpr}
 \usepackage{times}
 \usepackage{amsmath}
 \usepackage{amssymb}
\def\cvprPaperID{1}
\cvprfinalcopy
\usepackage{algorithmicx}
\usepackage[noend]{algpseudocode}
\usepackage{placeins}

\@ifundefined{showcaptionsetup}{}{%
 \PassOptionsToPackage{caption=false}{subfig}}
\usepackage{subfig}
\makeatother

\begin{document}

\title{Extraction and Classification of Diving Clips from Continuous Video
Footage}

\author{%
\begin{tabular}{cccc}
Aiden Nibali\textsuperscript{1} & Zhen He\textsuperscript{1} & Stuart Morgan\textsuperscript{1,2} & Daniel Greenwood\textsuperscript{2}\tabularnewline
\multicolumn{4}{c}{\textsuperscript{1} La Trobe University, \textsuperscript{2} Australian
Institute of Sport}\tabularnewline
\end{tabular}}
\maketitle
\begin{abstract}
Due to recent advances in technology, the recording and analysis of
video data has become an increasingly common component of athlete
training programmes. Today it is incredibly easy and affordable to
set up a fixed camera and record athletes in a wide range of sports,
such as diving, gymnastics, golf, tennis, etc. However, the manual
analysis of the obtained footage is a time-consuming task which involves
isolating actions of interest and categorizing them using domain-specific
knowledge. In order to automate this kind of task, three challenging
sub-problems are often encountered: 1) temporally cropping events/actions
of interest from continuous video; 2) tracking the object of interest;
and 3) classifying the events/actions of interest.

Most previous work has focused on solving just one of the above sub-problems
in isolation. In contrast, this paper provides a complete solution
to the overall action monitoring task in the context of a challenging
real-world exemplar. Specifically, we address the problem of diving
classification. This is a challenging problem since the person (diver)
of interest typically occupies fewer than 1\% of the pixels in each
frame. The model is required to learn the temporal boundaries of a
dive, even though other divers and bystanders may be in view. Finally,
the model must be sensitive to subtle changes in body pose over a
large number of frames to determine the classification code. We provide
effective solutions to each of the sub-problems which combine to provide
a highly functional solution to the task as a whole. The techniques
proposed can be easily generalized to video footage recorded from
other sports.
\end{abstract}

\section{Introduction}

Extracting useful information from video data has become more important
in recent years due to the increasing abundance of video data and
the low cost of data storage. Much research in this area is compartmentalized
into either solving action recognition and classification~\cite{scovanner_3-dimensional_2007,wang_action_2011,ji_3d_2013,ijjina_human_2016,yuan_action_2016},
where the algorithm predicts a discrete class label of actions, or
object tracking~\cite{mihaylova_object_????,comaniciu_real-time_2000,zhang_robust_2015},
where continuous pixel coordinates are predicted through time. However,
applications in the sports domain often require both problems to be
solved together in order to formulate a useful system. For example,
isolating individual goal attempts made by a row of training football
players in order to find problematic technique, or labeling the actions
performed by a gymnast when there are other people moving in the background,
or extracting and separating every forehand/backhand for one player
in a game of tennis. In each of these examples the person of interest
may only occupy a small region of the input frame, there are other
people not of interest within the frame, and an accurate understanding
of their actions requires an evaluation over an arbitrary temporal
span. This aspect of the problem in particular requires a novel approach
to learning. We refer to these types of problems as action monitoring
problems.

\begin{figure}
\begin{centering}
\noindent\begin{minipage}[t]{1\linewidth}%
\begin{center}
\includegraphics[width=1\linewidth]{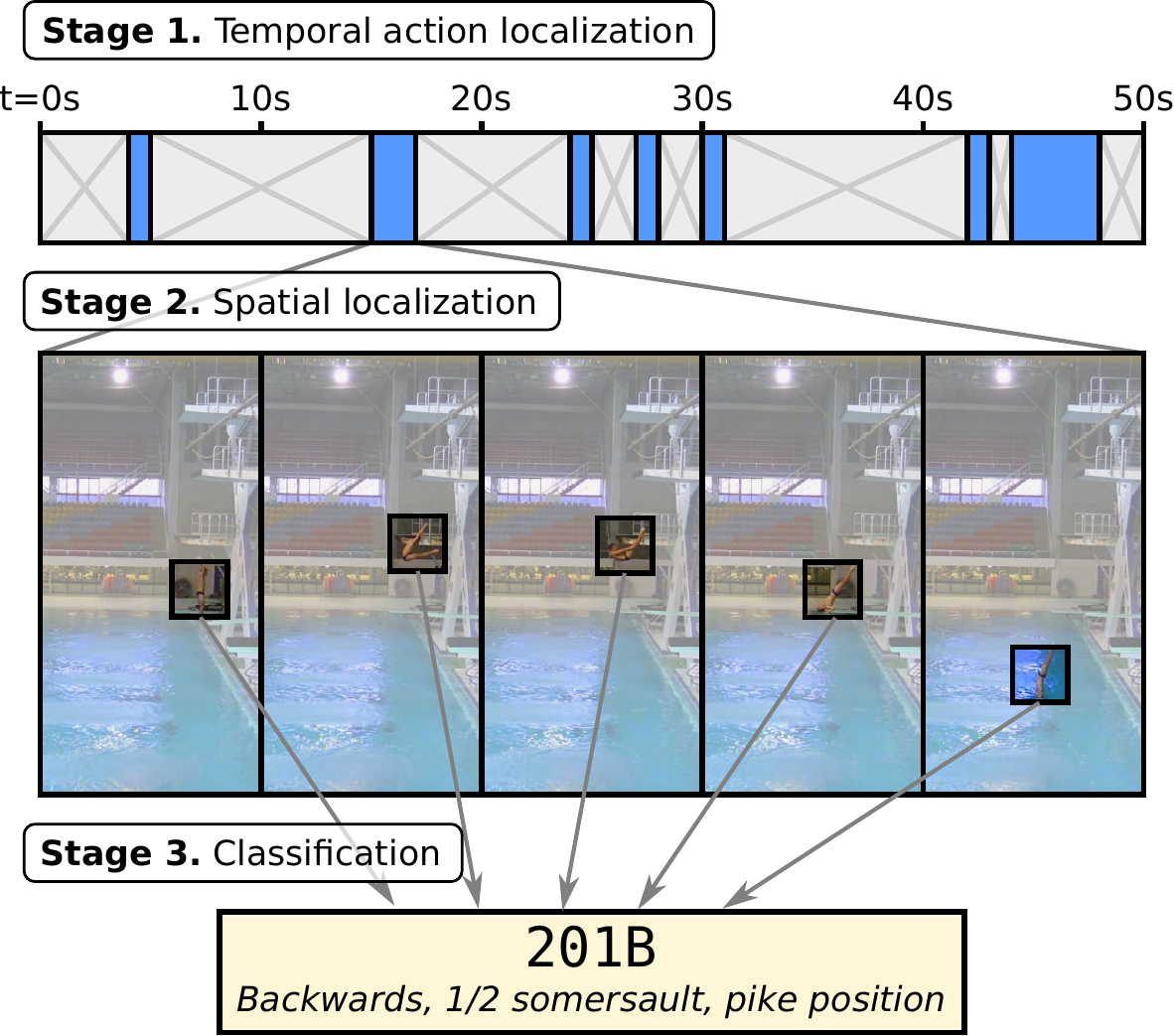}
\par\end{center}%
\end{minipage}
\par\end{centering}
\caption{\label{fig:activity-in-video} Our action clip extraction and classification
system. Each stage drills deeper into the data.}
\end{figure}

Solving the action monitoring problem requires solutions to the following
three sub-problems: 1) temporally cropping events/actions of interest
from continuous video; 2) tracking the person/animal of interest;
and 3) classifying the event/action of interest. Due to a lack of
publicly-available action monitoring data sets, this paper primarily
focuses on solving the diving monitoring problem using a novel data
set provided by the Australian Institute of Sport, as illustrated
in Figure \ref{fig:activity-in-video}. A solution requires that we
first identify the temporal bounds of each dive. We then track the
diver of interest to generate suitable spatial crops. Finally, we
need to feed the cropped images into a classifier. The solutions presented
have general application in the sports domain, and our approach can
be applied to solve many other action monitoring problems. The diving
monitoring problem is particularly hard since the diver occupies a
very small percentage of each frame (typically fewer than 1\% of the
pixels) and there are thousands of possible different dive codes.
So, using just a few pixels per frame we need to consistently separate
different dive types which differ only on subtle changes in diver
pose. In addition, the system needs to look at the entire dive (which
spans around 50-100 frames) to correctly assign a classification code.
In contrast, in most public video classification data sets the vast
majority of the classes can be assigned by just looking at 1 frame
of a video clip (\eg playing tennis versus playing basketball)~\cite{karpathy_large-scale_2014}.

We present a 3D convolutional neural network based solution for all
three sub-problems of temporal action localization, object tracking,
and action recognition. For temporal action localization, we predict
the probability that a frame is from the start, middle, and end of
a dive. This gives us higher confidence that a dive is correctly detected
since all three labels must be detected in sequence. The results show
we can correctly extract 98\% of dives, with a 26\% higher $F_{1}$
score than a straightforward baseline approach. For object tracking
we present a segmentation based solution to finding the center of
a diver in each frame. The results show our segmentation based solution
is appreciably more accurate than a more conventional regression based
solution. Finally, our proposed classification approach based on dilated
convolutions can achieve an average of 93\% accuracy for each component
of the dive codes.

\section{Related Work}

\paragraph{Video representation and classification}

At the heart of video analysis is the way the data is represented.
Many techniques extend 2D image representations to 3D by incorporating
the temporal dimension, including HOG3D~\cite{klaser_spatio-temporal_2008}
from HOG~\cite{dalal_histograms_2005}, extended SURF~\cite{willems_efficient_2008}
from SURF~\cite{bay_surf:_2006}, and 3D-SIFT~\cite{scovanner_3-dimensional_2007}
from SIFT~\cite{lowe_distinctive_????}. Other techniques such as
optical flow treat the temporal dimension as having properties distinct
from spatial dimensions. The work on dense trajectories proposed by
Wang \etal~\cite{wang_action_2011} takes such an approach, and
is currently a state-of-the-art hand-crafted feature algorithm for
video analysis. Unfortunately, the effectiveness of optical flow-based
techniques (including dense trajectories) comes at the price of computational
efficiency, which reduces their viability for real-time applications
and large-scale datasets.

Using learnt features via convolutional neural networks (CNNs) for
video analysis have become more popular since the huge success of
AlexNet~\cite{krizhevsky_imagenet_????} in the ILSVR 2012 image
classification challenge. One of the directions this research took
was in finding direct ways of applying 2D CNNs to video data by fusing
2D feature maps at different levels of the network hierarchy. Karpathy
\etal~\cite{karpathy_large-scale_2014} demonstrated that such fusion
schemes only achieve a modest improvement over using only a single
frame of input. Another direction taken was to treat video as 3D data
(with time being the 3rd dimension), and apply volumetric convolutions~\cite{tran_learning_2014}.
Such networks learn good representations of video data at the cost
of a large memory requirement.

There exist multiple more complex solutions for applying CNNs to action
recognition~\cite{simonyan2014two,yeung2015end,donahue2015long,yue2015beyond}.
Some of these solutions rely on optical flow~\cite{simonyan2014two,yue2015beyond},
which is slow to evaluate. Others rely on a recurrent architecture~\cite{yeung2015end,donahue2015long},
which is often difficult to train in practice.

\paragraph{Temporal action localization}

The dominant method for detecting the temporal extent of actions involves
sliding windows of several fixed lengths through the video, and classifying
each video segment to determine whether it contains an action~\cite{oneata2013action,weinzaepfel_learning_2015,shou_temporal_2016}.
The segment classifier can be based on hand-engineered feature descriptors~\cite{oneata2013action},
trained CNNs~\cite{shou_temporal_2016}, or a combination of the
two~\cite{weinzaepfel_learning_2015}. In contrast to this segment-based
approach, we are able to detect actions of arbitrary length without
sliding multiple windows through the video.

Other branches of work related to temporal action localization attempt
to solve different variations of the problem, such as detecting temporal
extents without explicit temporal annotations~\cite{lai2014recognizing,lai2014video,sun2015temporal},
or simultaneously detecting temporal and spatial boundaries~\cite{jain2014action,gkioxari2015finding}.

\paragraph{Object localization/detection}

Sermanet \etal~\cite{sermanet_overfeat:_2013} proposed a neural
network called OverFeat for object detection. OverFeat comprises of
a convolutional feature extractor and two network ``heads'' - one
for classification, and another for regression. The feature extractor
is similar to what is now commonly referred to as a fully-convolutional
network. This allows it to efficiently slide a window around the image
to extract features for different crops. The classifier is a multi-layer
perceptron which takes features from the feature extractor as input
and predicts a class as output. This tells us what is in each crop
(including confidence), and is already sufficient to produce coarse
bounding boxes. However, these boxes are refined further by training
a class-specific regression head which outputs bounding box dimensions
from image features.

Girshick \etal~\cite{girshick_rich_2014} proposed a different strategy
called R-CNN (regions with CNN features). They use an existing algorithm
(\eg Selective Search~\cite{uijlings_selective_2013} or EdgeBoxes~\cite{zitnick_edge_2014})
to produce region proposals, and warp the region of the image described
by each proposal to a fixed size. The warped image is run through
a CNN, the output features of which are used to prune the proposed
regions and generate final predictions. There now exist more efficient
works based on R-CNN which improve evaluation time~\cite{girshick_fast_2015,ren_faster_2015}.

Szegedy \etal~\cite{szegedy_deep_2013} proposed a segmentation-style
approach to object detection. Rather than dealing with region proposals
or output coordinates, the network takes the entire image as input
and produces a lower resolution ``mask'' depicting filled-in bounding
boxes at the output. The results reported in the paper are considerably
worse than R-CNN, but we note that this system is a more natural fit
for localization than detection due to complications introduced by
overlapping bounding boxes.

\paragraph{Tracking}

There has been a lot of research in the area of object tracking. In
this section we will focus on CNN based solutions~\cite{wang2015visual,li2016deeptrack,hong2015online,li2014robust,wang2013learning,wang2016stct,ma2015hierarchical}.
They all take the approach of tracking-by-detection, where a binary
classifier is applied to positive and negative samples from each frame.
Typically, the object bounding box of just the first frame is provided
and the CNN models learnt in an online manner. All methods need to
somehow deal with the small number of labeled training samples. \cite{wang2013learning}
pretrains the network using an autoencoder, \cite{wang2015visual,hong2015online,wang2016stct,ma2015hierarchical}
uses CNNs pretrained on the large ImageNet dataset and \cite{li2016deeptrack,li2014robust}
uses special loss functions and sampling techniques to cope with the
small number of training samples. In contrast to most existing work,
our solution first finds location candidates for each frame and then
applies global constraints to create the motion trajectory, which
is used to provide smoothly tracked output that aids the next stage
of the system.

\section{Overview}

At a high level, our dive detection and classification system consists
of three distinct stages, as shown in Figure \ref{fig:activity-in-video}.
Each stage uses a convolutional neural network at its core. Firstly,
we extract individual video clips of dives from continuous video footage.
Secondly, we localize the diver within each frame of the clip to produce
a tracking shot of the dive, which allows us to improve the ratio
of pixels in the clip which are useful for classification. Thirdly,
we use the tracked clip to predict the dive code using a classification
network.

The three stages of the system are linked by dependencies on preceding
stages. There are a few places where these dependencies led to different
design decisions from considering each stage in isolation. For instance,
during spatial object localization we fit the motion trajectory by
applying global constraints, and crop the images using a fixed-size
box to keep the scale consistent between frames. This smooth tracking
increases the accuracy of the classifier.

\section{Temporal action localization\label{sec:Temporal-action-localization}}

\begin{figure}
\begin{centering}
\begin{minipage}[t]{0.4\columnwidth}%
\begin{center}
\subfloat[\label{fig:single-probability-signal}Single probability signal]{\begin{centering}
\noindent\begin{minipage}[t]{1\linewidth}%
\begin{center}
\includegraphics[width=1\linewidth]{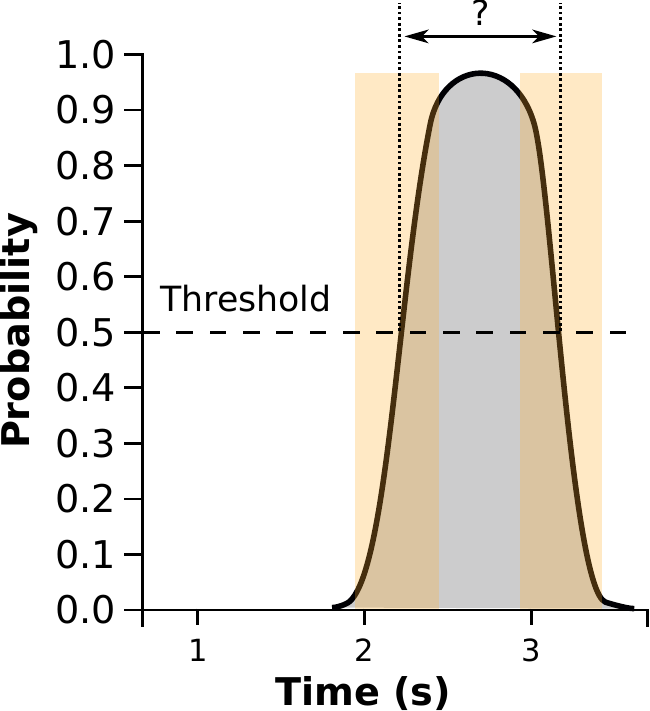}
\par\end{center}%
\end{minipage}
\par\end{centering}
}
\par\end{center}%
\end{minipage}\hspace*{\fill}%
\begin{minipage}[t]{0.4\columnwidth}%
\begin{center}
\subfloat[\label{fig:three-probability-signals}Three probability signals]{\begin{centering}
\noindent\begin{minipage}[t]{1\linewidth}%
\begin{center}
\includegraphics[width=1\linewidth]{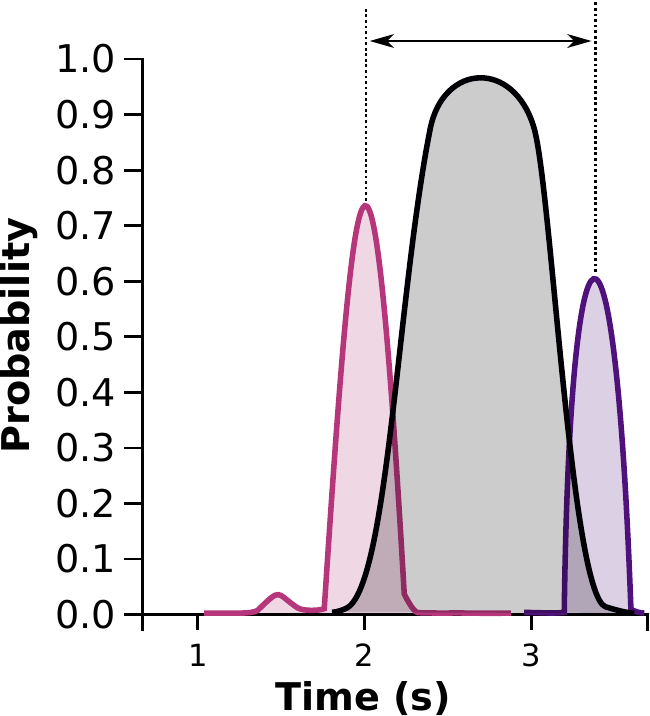}
\par\end{center}%
\end{minipage}
\par\end{centering}
}
\par\end{center}%
\end{minipage}\hspace*{\fill}%
\begin{minipage}[t][1\totalheight][b]{0.15\columnwidth}%
\begin{center}
\includegraphics[width=1\linewidth]{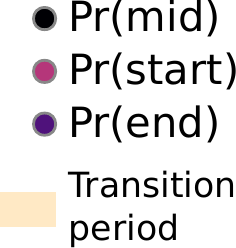}
\par\end{center}%
\end{minipage}
\par\end{centering}
\caption{\label{fig:dive-event-approaches} (a) It's difficult to tell precisely
when a dive starts and ends from the middle event probability only
due to the transition periods, (b) whereas the start and end probabilities
give more obvious time markers.}
\end{figure}

The first stage of the system involves extracting action clips from
continuous footage, a task known as \emph{temporal action localization}.
Our aim is to predict the temporal extent of each dive as accurately
as possible in order to crop the extracted clip tightly, thus maximizing
the number of frames which are relevant for classification. Hence
our network needs to be able to indicate the start and end times of
dives in a dynamic way. This differs from the existing temporal action
localization work with CNNs, which slide windows with one of several
predetermined lengths through the video~\cite{weinzaepfel_learning_2015,shou_temporal_2016}.

We explicitly identify three event states in the video footage: a
diver leaving a platform ($start$), a diver entering the water ($end$),
and any time during which the diver is airborne ($mid$). Our temporal
action localization neural network (TALNN) accepts 21 frames of video
as input, and outputs probabilities for the center frame containing
each of these events. These probabilities are predicted as independent
values (\ie they are not part of a single softmax), which allows
the network to output high probabilities for two events at once (\eg
start and middle). The network itself is built from volumetric convolutional
layers, with one head per probability signal. Table \ref{tab:talnn-arch}
specifies the architecture in detail. Each convolutional layer in
the body is followed by batch normalization and a ReLU non-linearity.

Let $\boldsymbol{x}_{t}$ be the 21-frame window centered at time
$t$. We now define the time varying probability signals $f_{M}(t)$
as in Equation \ref{eq:prob-signal}, where $M\in\{start,mid,end\}$.
\begin{equation}
f_{M}(t)=\Pr(M|\boldsymbol{x}{}_{t})\label{eq:prob-signal}
\end{equation}

At first it may seem unusual that we are considering the start and
end events at all, since the boundaries of the middle event should
be sufficient to determine when a dive starts and ends. Figure \ref{fig:single-probability-signal}
shows the problem with that approach. Namely we would need to select
some threshold (\eg 0.5) as to when the start and end boundaries
are defined. In contrast, Figure \ref{fig:three-probability-signals}
shows that using all three events (start, middle, and end) makes finding
the start and end of the dive less ambiguous. Note that we could theoretically
remove $f_{mid}(t)$ altogether, but we opt to keep it as a way of
reducing the likelihood of false positives.

\begin{figure}
\subfloat[\label{fig:raw-probability-signals}Raw probability signals]{\begin{centering}
\noindent\begin{minipage}[t]{1\columnwidth}%
\begin{center}
\includegraphics[width=1\linewidth]{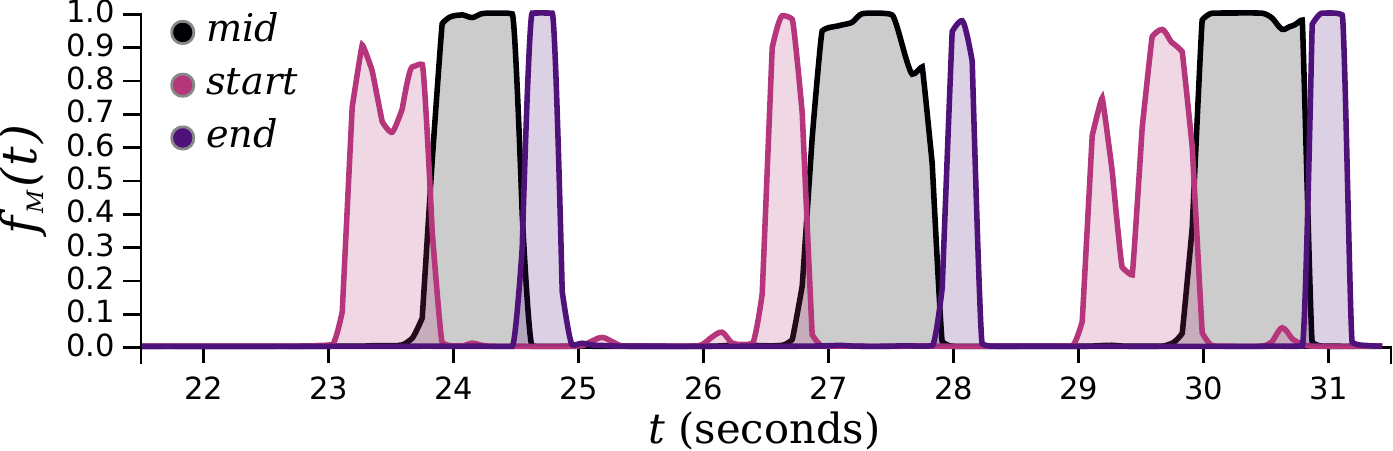}
\par\end{center}%
\end{minipage}
\par\end{centering}
}

\subfloat[\label{fig:smoothed-probability-signals}Smoothed probability signals]{\begin{centering}
\noindent\begin{minipage}[t]{1\columnwidth}%
\begin{center}
\includegraphics[width=1\linewidth]{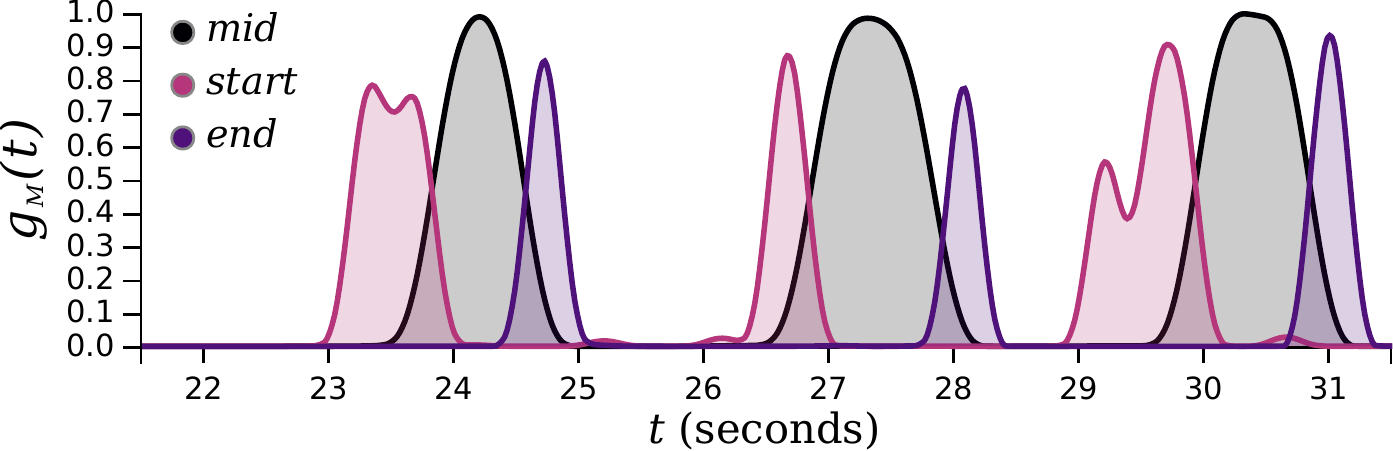}
\par\end{center}%
\end{minipage}
\par\end{centering}
}

\caption{\label{fig:dive-event-probs} Raw and smoothed probability signals
for a section of video footage containing three dives.}
\end{figure}

After training the TALNN to identify the different types of events,
$f_{M}(t)$ is obtained by sliding a 21-frame window through the video
and evaluating the network. Figure \ref{fig:raw-probability-signals}
shows that although the output provides a strong indication of when
dives occur, it is not perfectly smooth.

\paragraph{Smoothing}

To make the peaks in the probability signals more pronounced we process
them further into smoothed probability signals, $g_{M}(t)$ (Figure
\ref{fig:smoothed-probability-signals}). This makes the final dive
extraction step more robust. A common way of smoothing signals is
to apply a window function, as in Equation \ref{eq:applying-window}.

\begin{equation}
g_{M}(t)=\frac{\intop_{-\infty}^{\infty}f_{M}(\tau)w(\tau-t+T/2)d\tau}{\intop_{-\infty}^{\infty}w(\tau)d\tau}\label{eq:applying-window}
\end{equation}

We use the Hann window function (Equation \ref{eq:hann}) for smoothing,
which gives us the formula for calculating $g_{M}(t)$ described in
Equation \ref{eq:hann-smoothing}.

\begin{equation}
w(t)=\begin{cases}
\sin^{2}\left(\frac{\pi t}{T}\right) & \text{{if\;}}0\le t\le T\\
0 & \text{{otherwise}}
\end{cases}\label{eq:hann}
\end{equation}

\begin{equation}
g_{M}(t)=\frac{2}{T}\intop_{t-T/2}^{t+T/2}f_{M}(\tau)\sin^{2}\left(\frac{\pi(\tau-t)}{T}+\frac{\pi}{2}\right)d\tau\label{eq:hann-smoothing}
\end{equation}

\paragraph{Extraction}

Given the three smoothed probability signals, we can apply a simple
algorithm to extract concrete dive intervals. Firstly, identify candidate
dives by locating peaks in $g_{mid}(t)$. Secondly, perform a limited
scan forwards and backwards through time (we use 1 second) to locate
the dive's start and end from peaks in their respective probability
signals. If there are no strong nearby peaks in the start and end
probability signals, discard the dive candidate.

\begin{table}
\begin{centering}
\begin{minipage}[t]{0.47\columnwidth}%
\begin{center}
\subfloat[\label{tab:talnn-arch}Temporal localization]{\begin{centering}
\noindent\begin{minipage}[t]{1\linewidth}%
\begin{center}
{\footnotesize{}}%
\begin{tabular}{|c|c}
\hline 
{\footnotesize{}Body} & \multicolumn{1}{c|}{{\footnotesize{}Head}}\tabularnewline
\hline 
\hline 
{\footnotesize{}conv3-32, strided} & \multicolumn{1}{c|}{{\footnotesize{}conv1-1}}\tabularnewline
\hline 
{\footnotesize{}conv3-32} & \multicolumn{1}{c|}{{\footnotesize{}avgpool}}\tabularnewline
\hline 
{\footnotesize{}conv3-64, strided} & \multicolumn{1}{c|}{{\footnotesize{}sigmoid}}\tabularnewline
\hline 
{\footnotesize{}conv3-64} & \multirow{5}{*}{}\tabularnewline
\cline{1-1} 
{\footnotesize{}conv3-128, strided} & \tabularnewline
\cline{1-1} 
{\footnotesize{}conv3-128} & \tabularnewline
\cline{1-1} 
{\footnotesize{}conv3-256, strided} & \tabularnewline
\cline{1-1} 
{\footnotesize{}conv3-256} & \tabularnewline
\cline{1-1} 
\end{tabular}
\par\end{center}%
\end{minipage}
\par\end{centering}
}
\par\end{center}%
\end{minipage}\hspace*{\fill}%
\begin{minipage}[t]{0.45\columnwidth}%
\begin{center}
\subfloat[\label{fig:localization-arch}Spatial localization context net]{\begin{centering}
\noindent\begin{minipage}[t]{1\linewidth}%
\begin{center}
{\footnotesize{}}%
\begin{tabular}{|c|c|c|}
\hline 
{\footnotesize{}Kernel} & {\footnotesize{}Dilation} & {\footnotesize{}Maps}\tabularnewline
\hline 
\hline 
{\footnotesize{}3x3x3} & {\footnotesize{}1x1x1} & {\footnotesize{}2}\tabularnewline
\hline 
{\footnotesize{}3x3x3} & {\footnotesize{}1x1x1} & {\footnotesize{}2}\tabularnewline
\hline 
{\footnotesize{}3x3x3} & {\footnotesize{}2x2x2} & {\footnotesize{}4}\tabularnewline
\hline 
{\footnotesize{}3x3x3} & {\footnotesize{}4x4x4} & {\footnotesize{}4}\tabularnewline
\hline 
{\footnotesize{}3x3x3} & {\footnotesize{}8x8x8} & {\footnotesize{}4/8}\tabularnewline
\hline 
{\footnotesize{}3x3x3} & {\footnotesize{}1x16x16} & {\footnotesize{}8/16}\tabularnewline
\hline 
{\footnotesize{}3x3x3} & {\footnotesize{}1x1x1} & {\footnotesize{}8/16}\tabularnewline
\hline 
{\footnotesize{}1x1x1} & {\footnotesize{}1x1x1} & {\footnotesize{}1/3}\tabularnewline
\hline 
\end{tabular}
\par\end{center}%
\end{minipage}
\par\end{centering}
}
\par\end{center}%
\end{minipage}
\par\end{centering}
\caption{CNN localization architectures.}
\end{table}

\section{Spatial localization}

The aim of the spatial localization stage is to produce a trajectory
consisting of the diver centroid in each frame of the input clip.
Given a list of centroids we can then take a fixed size crop from
each frame to produce a tracking clip, which will supply the classifier
with fixed size input of a consistent scale that excludes most of
the background.

We take a tracking-by-detection approach to spatial localization,
which is separated into two steps. The first step is to find object
location candidates which indicate potential locations for the diver
in each frame. The second step is to take these candidate locations
and apply global constraints to construct a motion trajectory.

\subsection{Object location candidates}

Here we compare three possible solutions to the object location candidate
proposal step which we refer to as full regression, partial regression,
and segmentation.

\paragraph{Full regression}

Perhaps the most straightforward approach to spatial localization
is to take a complete video clip as input, and attempt to train a
network which outputs the object location coordinates $\left(l_{x},l_{y}\right)$
directly for each frame. We call this approach “full regression”.
One advantage of full regression is that it gives a single location
per frame, which removes the need for a second step to construct the
motion trajectory. In practice we found full regression to yield very
poor accuracy of predicted locations, with a high amount of location
``jitter'' between neighboring frames.

\paragraph{Partial regression}

An alternative to full regression is to only consider a small crop
of the input at a time (an ``input patch''), and train a network
to predict whether the object is contained in the patch. The network
is also trained to output its location relative to the input patch's
frame of reference, though patches which do not contain the object
exclude the location from loss calculations. This is an approach used
successfully in prominent object detection systems including OverFeat~\cite{sermanet_overfeat:_2013}
and Fast R-CNN~\cite{girshick_fast_2015}.

The network used in this paper for partial regression is a stack of
two context networks~\cite{yu_multi-scale_2015} followed by an average
pooling layer. Table \ref{fig:localization-arch} specifies our configuration
for the context networks. The use of dilated convolutions improves
the scale invariance of the network, which helps with the fact that
divers are at different distances from the camera. Furthermore, we
were able to construct the network with very few feature maps, resulting
in a compact model.

An important aspect of our implementation of partial regression is
that the network is fully convolutional. This means that at inference
time we can provide the entire image as input (rather than patches).
The network will then implicitly slide a window through the image,
but do so in a way which shares common intermediate activations. This
is much more efficient than explicitly making overlapping crops and
feeding them through the network separately. The overlap of the windows
can be adjusted by altering the stride of the average pooling layer.

\paragraph{Segmentation}

\begin{figure}
\begin{centering}
\begin{minipage}[t]{0.3\columnwidth}%
\begin{center}
\subfloat[\label{fig:segmentation_input}Input]{\begin{centering}
\noindent\begin{minipage}[t]{1\linewidth}%
\begin{center}
\includegraphics[width=1\linewidth]{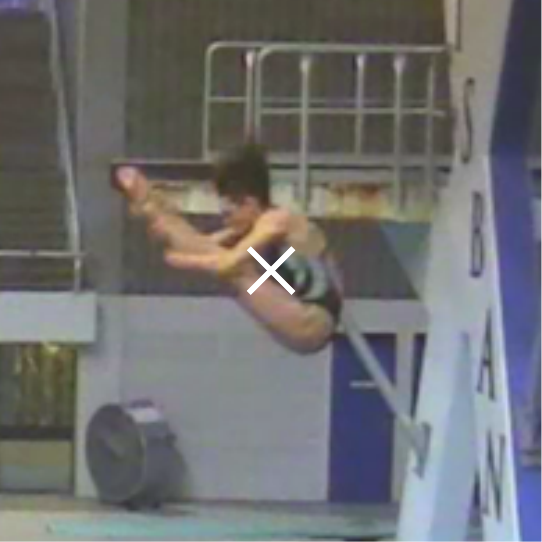}
\par\end{center}%
\end{minipage}
\par\end{centering}
}
\par\end{center}%
\end{minipage}\hspace*{\fill}%
\begin{minipage}[t]{0.3\columnwidth}%
\begin{center}
\subfloat[\label{fig:segmentation_target}Target]{\begin{centering}
\noindent\begin{minipage}[t]{1\linewidth}%
\begin{center}
\includegraphics[width=1\linewidth]{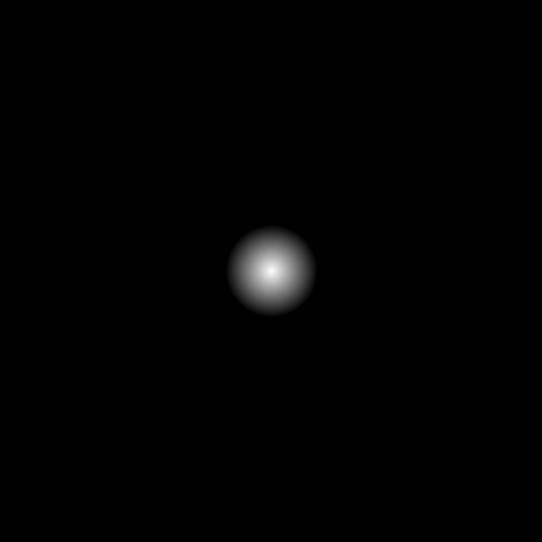}
\par\end{center}%
\end{minipage}
\par\end{centering}
}
\par\end{center}%
\end{minipage}\hspace*{\fill}%
\begin{minipage}[t]{0.3\columnwidth}%
\begin{center}
\subfloat[\label{fig:segmentation_output}Output]{\begin{centering}
\noindent\begin{minipage}[t]{1\linewidth}%
\begin{center}
\includegraphics[width=1\linewidth]{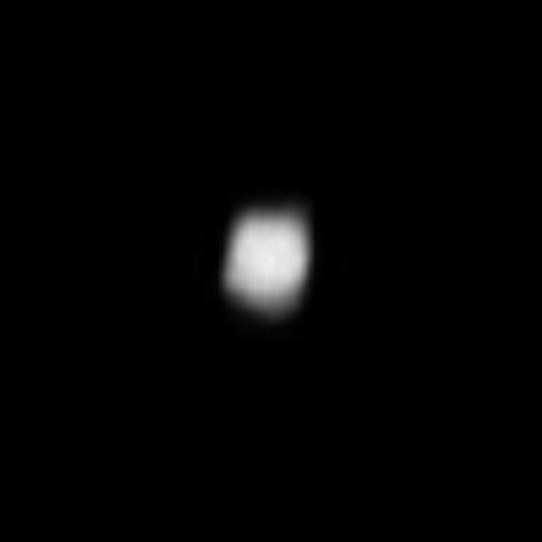}
\par\end{center}%
\end{minipage}
\par\end{centering}
}
\par\end{center}%
\end{minipage}
\par\end{centering}
\caption{\label{fig:hot-spot}Using a ``hot-spot'' for localization via segmentation.
Diver location is marked on input for reference only.}
\end{figure}

We observe that going from full to partial regression resulted in
much more accurate location candidates, and that a key difference
is that the latter places less emphasis on regressing coordinates.
We decided to take a step further in this direction and eliminate
regression completely, which is achieved by reframing the problem
as a segmentation problem. Instead of using numeric coordinates as
the target, we artificially generate target images for each frame
where the location of the diver is indicated with a fixed-size ``hot-spot''
(Figure \ref{fig:segmentation_target}). The network learns to output
blob-like approximations of these hot-spots (Figure \ref{fig:segmentation_output})
which can then be converted into centroids using existing techniques
for blob detection~\cite{opencv_blob}. The main advantage of this
approach is that it unburdens the network of transforming spatial
activations into numeric coordinates.

The segmentation style of temporal localization has an imbalance in
the output, as the hot-spot occupies a small portion of the patch.
With a traditional loss function like binary cross-entropy (BCE),
this makes the prediction of all zeros an attractive behavior for
the network to learn in terms of loss minimization. To counteract
this, we modified BCE to weight positive outputs higher, thus penalizing
the network more harshly for ignoring them (Equation \ref{eq:weighted_bce}).
\begin{equation}
\mathcal{L}=\frac{-\log(\hat{y})y}{2(1-\beta)}+\frac{-\log(1-\hat{y})(1-y)}{2\beta}\label{eq:weighted_bce}
\end{equation}

\begin{figure}
\begin{centering}
\noindent\begin{minipage}[t]{1\linewidth}%
\begin{center}
\includegraphics[width=1\linewidth]{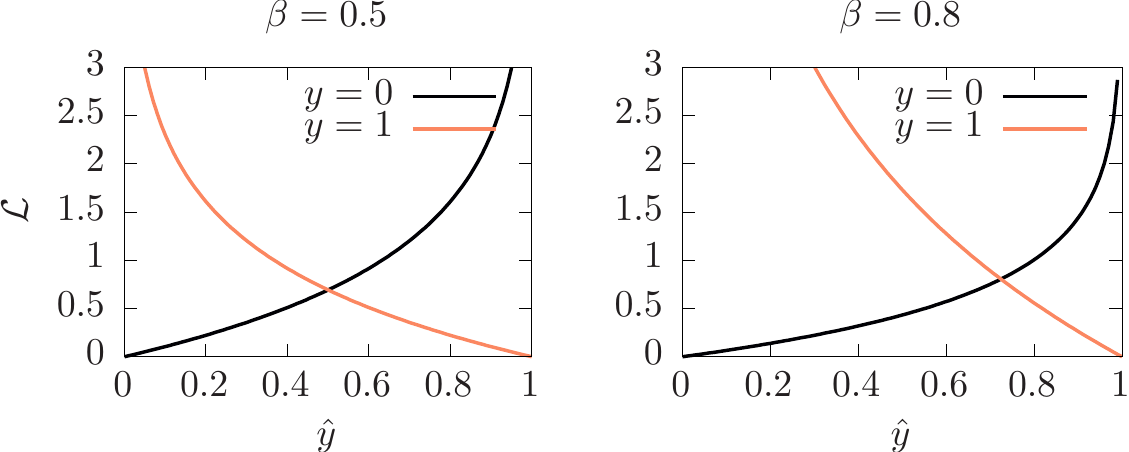}
\par\end{center}%
\end{minipage}
\par\end{centering}
\caption{\label{fig:weighted-bce}Side-by-side comparison of standard BCE (left)
and weighted BCE (right). }
\end{figure}

When $\beta=0.5$, weighted BCE is equivalent to the usual BCE formulation.
When $\beta\in(0.5,1)$, the positive example term of the loss function
is weighted higher. Figure \ref{fig:weighted-bce} illustrates how
weighted BCE imposes a greater loss for misclassified positive examples
than negative ones when $\beta>0.5$. We found $\beta=0.8$ to work
well in practice.

As with partial regression, we train the segmentation network on input
patches. The network architecture is similar, the main difference
being that the average pooling layer is removed and second context
network adjusted such that there is 1 output per pixel.

\subsection{Global constraints}

Neither the partial regression nor the segmentation approach is able
to produce a proper motion trajectory alone, as there can be many
(or zero) locations output for each frame. We get around this by using
a second step which applies global constraints to refine the location
candidates and produce a motion trajectory. Ultimately this produces
one location per frame to center the crop on when constructing a tracking
clip. During this step bad location candidates are rejected and missing
locations are interpolated.

The appropriate constraints to apply when constructing a motion trajectory
will depend on the problem. For diving, we have the ability to apply
very strong constraints derived from basic kinematic formulae. In
fact, we can go so far as to specify a model for the trajectory which
has only five parameters – two for a linear mapping from time to horizontal
location, and three for a quadratic mapping from horizontal location
to vertical location. Algorithm \ref{alg:create-ransac-model} describes
how the model is constructed. Once we have a known model we can use
the RANSAC~\cite{fischler_random_1981} algorithm to find the instance
of the model which best fits the location candidates. RANSAC is an
iterative algorithm which fits the data by repeatedly creating model
instances for random subsets of points and selecting whichever one
fits the complete set of points best. The main benefit of using RANSAC
is that it is very robust to contamination from outliers, and is therefore
able to ignore bad location candidates. In practice we used the improved
MSAC~\cite{torr_mlesac:_2000} variant of RANSAC which generally
fits the model in fewer iterations.

\begin{algorithm}
\begin{algorithmic}[0]
\small
\Function{CreateModel}{ts[], xs[], ys[]}
\State $a_0, a_1 \gets LinearRegression(ts, xs)$
\State $b_0, b_1, b_2 \gets QuadraticRegression(xs, ys)$
\Function{Model}{t}
\State $x \gets a_0 + a_1t$
\State $y \gets b_0 + b_1x + b_2x^2$
\State \Return $x, y$
\EndFunction
\State \Return Model
\EndFunction
\end{algorithmic}

\caption{\label{alg:create-ransac-model}Creating a motion trajectory model.}
\end{algorithm}

For less tightly constrained problems, an alternative method for constructing
the motion trajectory must be employed. Although we did not explore
this space ourselves, one approach would be to use local feature descriptors
to track candidate locations through time.

\section{Classification\label{sec:Classification}}

\begin{figure}
\begin{centering}
\noindent\begin{minipage}[t]{1\linewidth}%
\begin{center}
\includegraphics[width=0.9\linewidth]{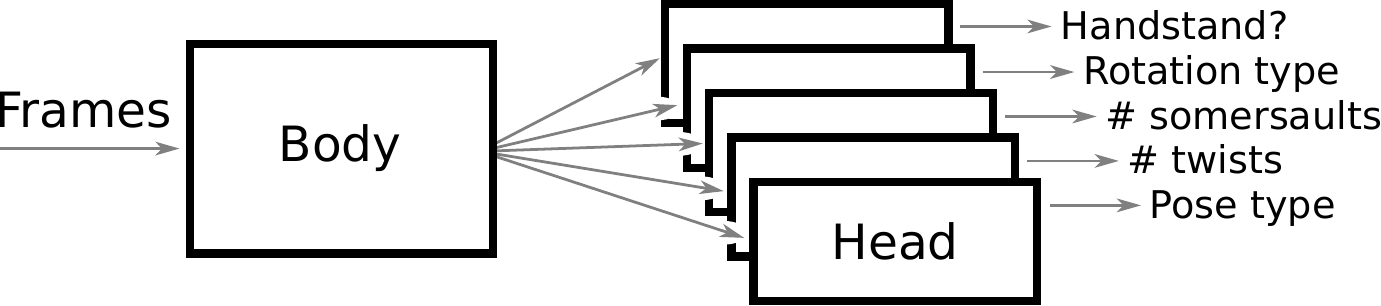}
\par\end{center}%
\end{minipage}
\par\end{centering}
\caption{\label{fig:classifier-arch}High-level view of the classifier architecture
with a head for each part of the dive code. }
\end{figure}

Classifying dives involves outputting a five-part code, where each
part represents a different property of the dive. An example of a
dive code is \texttt{201B}, where the 2 means backwards rotation,
the 1 means one half-somersault, the B means pike position, and the
overall code implies that there are no twists and no handstand start.
We could try to classify the entire code using a single output representation,
which equates to a 1-in-$k$ classification problem where $k$ is
the number of combinations of all properties. This would result in
thousands of possible output classes, most of which would have just
a few or zero training examples.

Instead, we propose using multi-task learning consisting of a single
network with 5 heads, each outputting a separate property (Figure
\ref{fig:classifier-arch}). One way to reason about the architecture
is that the network body learns to extract features from the input
which are relevant for predicting one or more parts of the dive code.
The heads take these features and use them to predict a particular
part of the dive code. Our hypothesis is that some intermediate features
can be shared between heads, making it easier for the network to rule
out unlikely dive code combinations. We use a deep convolutional network
for the model body, and multi-layer perceptrons for the heads. The
internal structure of each head is identical, except for the number
of outputs.

The classification network takes tracked video clips as input. Since
the clips are now cropped around the diver, we can use a higher resolution
than the previous networks under the same memory constraints. To keep
the input size constant we always temporally downsample the clip to
a length of 16 frames, which we verified is sufficient to solve the
classification task as a human annotator.

\begin{table}
\begin{centering}
\begin{tabular}{|c|c|c|}
\hline 
{\footnotesize{}C3D} & {\footnotesize{}C3D (alt.)} & {\footnotesize{}Dilated}\tabularnewline
\hline 
\hline 
\multicolumn{3}{|c|}{\emph{\footnotesize{}Body}}\tabularnewline
\hline 
{\footnotesize{}conv3-64} & {\footnotesize{}conv3-32, BN} & {\footnotesize{}conv3-32, BN}\tabularnewline
\hline 
{\footnotesize{}1x2x2 maxpool} & {\footnotesize{}1x2x2 maxpool} & {\footnotesize{}1x2x2 maxpool}\tabularnewline
\hline 
{\footnotesize{}conv3-128} & {\footnotesize{}conv3-64, BN} & {\footnotesize{}conv3-64, BN}\tabularnewline
\hline 
{\footnotesize{}2x2x2 maxpool} & {\footnotesize{}2x2x2 maxpool} & {\footnotesize{}2x2x2 maxpool}\tabularnewline
\hline 
{\footnotesize{}conv3-256 ($\times2$)} & {\footnotesize{}conv3-128, BN ($\times2$)} & {\footnotesize{}conv3-128, BN ($\times2$)}\tabularnewline
\hline 
{\footnotesize{}2x2x2 maxpool} & {\footnotesize{}2x2x2 maxpool} & {\footnotesize{}2x2x2 maxpool}\tabularnewline
\hline 
{\footnotesize{}conv3-512 ($\times2$)} & {\footnotesize{}conv3-256, BN ($\times2$)} & {\footnotesize{}conv3-256, BN ($\times2$)}\tabularnewline
\hline 
{\footnotesize{}2x2x2 maxpool} & {\footnotesize{}2x2x2 maxpool} & {\footnotesize{}-}\tabularnewline
\hline 
{\footnotesize{}conv3-512 ($\times2$)} & {\footnotesize{}conv3-256, BN ($\times2$)} & {\footnotesize{}conv3-d2-256, BN ($\times2$)}\tabularnewline
\hline 
{\footnotesize{}2x2x2 maxpool} & {\footnotesize{}2x2x2 maxpool} & {\footnotesize{}-}\tabularnewline
\hline 
{\footnotesize{}-} & {\footnotesize{}dropout-0.5} & {\footnotesize{}dropout-0.5}\tabularnewline
\hline 
\multicolumn{3}{|c|}{\emph{\footnotesize{}Head}}\tabularnewline
\hline 
{\footnotesize{}fc-4096} & {\footnotesize{}fc-2048, BN} & {\footnotesize{}conv1-12}\tabularnewline
\hline 
{\footnotesize{}dropout-0.5} & {\footnotesize{}-} & {\footnotesize{}context net}\tabularnewline
\hline 
{\footnotesize{}fc-4096} & {\footnotesize{}fc-2048, BN} & {\footnotesize{}2x2x2 maxpool}\tabularnewline
\hline 
{\footnotesize{}dropout-0.5} & {\footnotesize{}-} & {\footnotesize{}conv3-12, BN}\tabularnewline
\hline 
{\footnotesize{}fc-output} & {\footnotesize{}fc-output} & {\footnotesize{}conv3-output, avgpool}\tabularnewline
\hline 
\end{tabular}
\par\end{centering}
\caption{\label{tab:c3d-archs}Architectural differences between vanilla C3D
and our variations used for classification.}
\end{table}

In this paper we consider three classifier architectures (Table \ref{tab:c3d-archs}),
all of which are based on the ``C3D'' volumetric convolutional network
proposed by Tran \etal~\cite{tran_learning_2014}. The first is
a direct implementation of the C3D architecture which follows the
original work closely. All layers up until and excluding the first
fully connected layer form the model body, and the rest form a head.
The second is an altered version of C3D which makes room for batch
normalization (BN)~\cite{ioffe_batch_2015} by halving the number
of features throughout the network. The third architecture introduces
dilated convolutions for scale invariance~\cite{yu_multi-scale_2015}.
Pooling in the latter half of the body is removed, and the last two
convolutional layers given a dilation of 2 (conv3-d2) to maintain
receptive field size. A context network~\cite{yu_multi-scale_2015}
with layers 5 and 6 removed is introduced into the head for multi-scale
aggregation.

\section{Data set}

The data set consists of 25 hours of video footage containing 4716
non-overlapping sport dives. The video was recorded over 10 days of
athlete training at the Brisbane Aquatic Centre. The scene is observed
from the perspective of a fixed camera which has 9 platforms and springboards
at varying heights and distances in view. Each dive is labeled with
a start and end time, along with a code representing the type of dive
performed. 20\% of the dives are also labeled with a quadratic curve
describing the location of the athlete in each frame of the dive.
An additional day's worth of footage containing 612 dives is kept
aside as the test set.

The dive code encodes 5 distinct properties of the dive: rotation
type, pose type, number of somersaults, number of twists, and whether
the dive began with a handstand. These properties are not all represented
uniformly in the data set. For instance, dives involving twists are
uncommon, and dives starting with a handstand are even rarer.

\section{Experiments}

\subsection{Temporal action localization}

\begin{figure}
\begin{centering}
\begin{minipage}[t][1\totalheight][c]{0.7\linewidth}%
\begin{center}
\includegraphics[width=0.8\linewidth]{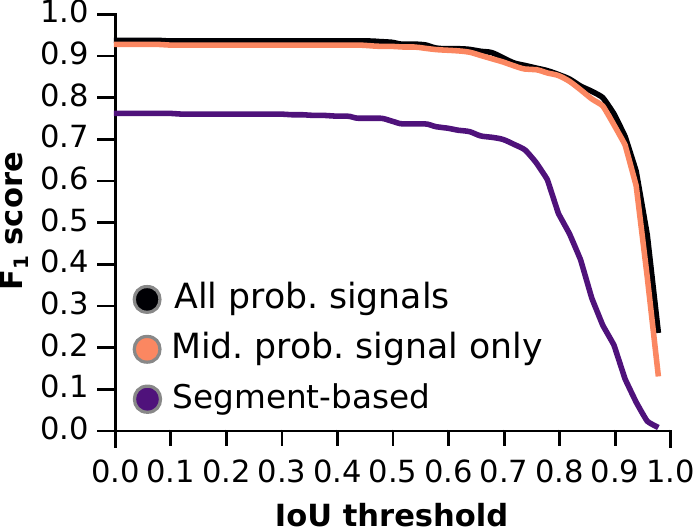}
\par\end{center}%
\end{minipage}
\par\end{centering}
\caption{\label{fig:fscore-vs-iou-thresh}Temporal action localization results
as the IoU threshold was varied.}
\end{figure}

As a point of comparison we implement a segment-based temporal action
localization method based on the work of Shou \etal~\cite{shou_temporal_2016}.
We use a single C3D-based network to directly predict how well a particular
segment matches any sort of dive. We incorporate batch normalization
into the network in the same way as the classification network, and
do not perform any pretraining. Although we did not explicitly gather
time metrics, we will note that performing inference on multiple segment
lengths did make the segment-based system very slow to evaluate.

With the segment-based approach established as a baseline, we consider
two of our own approaches to temporal action localization as discussed
in Section \ref{sec:Temporal-action-localization}. The first approach
uses only a single probability signal indicating the middle of a dive
(Figure \ref{fig:single-probability-signal}), with the transition
threshold set to 0.5. The second approach uses three probability signals
for the start, middle, and end (Figure \ref{fig:three-probability-signals}).
Each network was trained to convergence using ADADELTA~\cite{zeiler_adadelta:_2012}.
A predicted dive interval is deemed ``correct'' if it matches a
labeled dive interval with an IoU (intersection over union) above
a certain threshold. A ``false positive'' is a predicted interval
without a corresponding labeled dive, and a ``false negative'' is
a labeled dive not predicted by the system.

Figure \ref{fig:fscore-vs-iou-thresh} shows a plot of the $F_{1}$
score for the different approaches as the IoU threshold was varied.
Both of our approaches (all probability signals and middle probability
signal only) perform much better than the segment-based approach,
which is unable to reach an $F_{1}$ score of 0.8 for any IoU threshold.
Although the performance of our own two approaches are similar, we
advocate using all three signals since doing so shows slightly better
results, and in other situations it may be more difficult to threshold
the middle probability signal.

\begin{table}
\begin{centering}
\begin{tabular}{|c|c|c|c|}
\hline 
 & {\footnotesize{}Precision} & {\footnotesize{}Recall} & {\footnotesize{}$F_{1}$ score}\tabularnewline
\hline 
\hline 
{\footnotesize{}Segment-based~\cite{shou_temporal_2016}} & {\footnotesize{}0.7671} & {\footnotesize{}0.7157} & {\footnotesize{}0.7405}\tabularnewline
\hline 
{\footnotesize{}Ours} & \textbf{\footnotesize{}0.8825} & \textbf{\footnotesize{}0.9829} & \textbf{\footnotesize{}0.9296}\tabularnewline
\hline 
\end{tabular}
\par\end{centering}
\caption{\label{tab:TALNN-results}Action clip extraction results.}
\end{table}

Table \ref{tab:TALNN-results} shows the precision, accuracy, and
$F_{1}$ score for our three-signal approach and the segment-based
approach, with the IoU threshold set to 0.5. At first it seems as
if the precision of the TALNN is much worse than its recall. However,
upon examining the false positives it was found that the vast majority
did in fact contain dives that were simply not labeled in the data
set. During our manual inspection of the false positives we did not
find a single example that wasn't a labeling mistake. On the other
hand, dives which were missed by the TALNN were mostly legitimate
oversights, with dives from the furthest springboard being the most
common culprit.

The results of the TALNN stage are very convincing, and provide a
solid starting point for the rest of the system. The segment-based
approach does not achieve performance metrics which are as strong.
We believe that the main reason for this is the fixed segment lengths
– any dive which does not perfectly match a segment length will inevitably
incur error from the difference.

\subsection{Spatial localization}

\begin{figure}
\begin{centering}
\noindent\begin{minipage}[t][1\totalheight][c]{1\linewidth}%
\begin{center}
\includegraphics[width=1\linewidth]{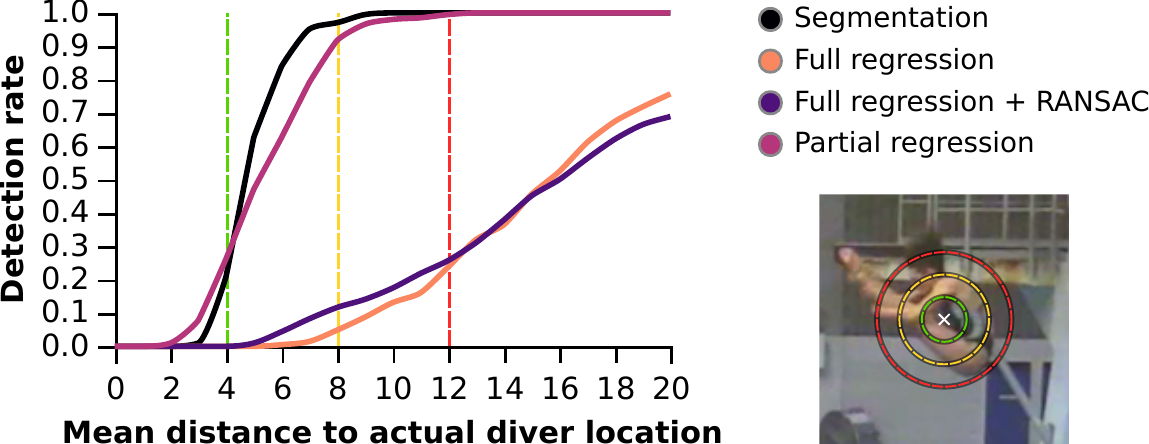}
\par\end{center}%
\end{minipage}
\par\end{centering}
\caption{\label{fig:locator-results}Spatial localization network results.
Three distances are marked on a video frame for reference.}
\end{figure}

Each network was trained to convergence using ADADELTA~\cite{zeiler_adadelta:_2012}.
The hot-spot to location conversion for the segmentation approach
was handled using OpenCV's blob detector~\cite{opencv_blob}, which
leverages the contour finding algorithms proposed by Suzuki \etal~\cite{suzuki_topological_1985}.

Figure \ref{fig:locator-results} shows, for a range of distance error
thresholds, the percentage of dive clips that had a mean error distance
below that threshold. Closer to the top-left is better, as this indicates
high detection rate within a strict distance limit. The results show
just how poorly the full regression approach performs, even when global
constraints are applied using RANSAC. Upon inspecting individual examples,
we found that the full regression network often seemed to ignore subtleties
of the current dive instance in favor of some learnt statistical average
across the training set location labels. The margin between the partial
regression approach and our novel segmentation approach is less pronounced,
but shows that segmentation does indeed work best.

In practice we found that partial regression resulted in many more
candidate locations than segmentation. This was not an issue for RANSAC
due to its speed and robustness to contamination, but we note that
other techniques for constructing motion trajectories may benefit
heavily from the reduced number of candidates produced by segmentation.

\subsection{Classification}

\begin{figure}
\begin{centering}
\noindent\begin{minipage}[t]{1\linewidth}%
\begin{center}
\includegraphics[width=1\linewidth]{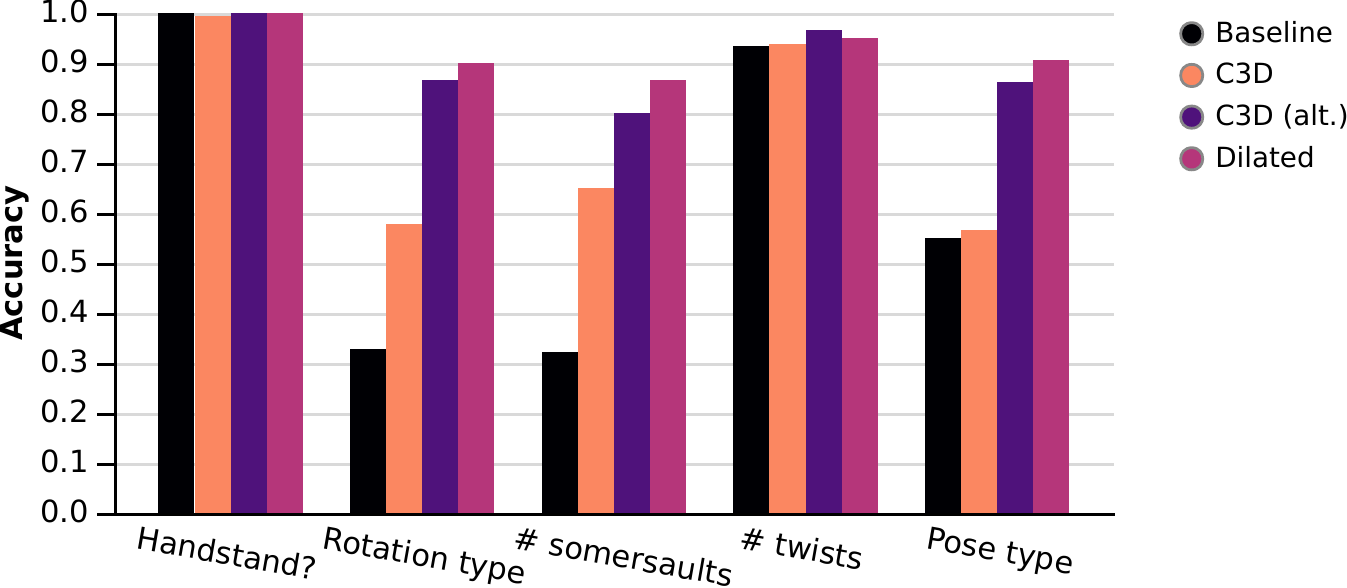}
\par\end{center}%
\end{minipage}
\par\end{centering}
\caption{\label{fig:classifier-results}Classifier network accuracy.}
\end{figure}

Since the data set does not contain an equal number of examples for
each type of dive, we include a baseline to help visualize this skew.
The baseline shows the results of always outputting the statistical
mode for each part of the dive code. Gains in accuracy above this
baseline are indicative of the system's ability to discriminate between
classes.

Table \ref{tab:c3d-archs} specified the architecture of each classifier
model. The networks make heavy use of volumetric convolutions with
$3\times3\times3$ kernels and use ReLU non-linearities. Regularization
is provided by dropout~\cite{hinton_improving_2012} and, for two
of the architectures, batch normalization. Each network was trained
until convergence using stochastic gradient descent with a momentum
of 0.9 and an initial learning rate of 0.006 (0.003 for vanilla C3D),
which is halved every 30 epochs.

Figure \ref{fig:classifier-results} shows accuracy results for the
classification networks when isolated from the other stages (\ie
using ground truth labels for diver locations). Despite halving the
number of feature maps in order to fit batch normalization, we observe
that doing so still leads to a marked improvement in accuracy. We
suspect that the increased regularization provided by batch normalization
is contributing a lot to the performance of the network, as our data
set is relatively small in comparison to existing large-scale public
image data sets.

Adding dilated convolutions to the altered C3D network resulted in
a boost to classification accuracy for all parts of the dive code
except the twist count. We theorize that the dilations increase the
network's ability to recognize features irrespective of the distance
of the diver from the camera.

\subsubsection{Combined classification}

In order to measure the impact that errors introduced in the temporal
and spatial localization stages have on classification, we conducted
a combined classification experiment using three-signal temporal action
localization, localization by segmentation, and classification with
dilations. Table \ref{tab:end-to-end-results} shows that although
error from the earlier stages does have a negative impact on classification
accuracy, the complete system is still viable.

\begin{table}
\begin{centering}
\begin{tabular}{|c|c|c|}
\hline 
 & {\footnotesize{}Isolated} & {\footnotesize{}Combined}\tabularnewline
\hline 
\hline 
{\footnotesize{}Handstand?} & {\footnotesize{}100.00\%} & {\footnotesize{}99.67\%}\tabularnewline
\hline 
{\footnotesize{}Rotation type} & {\footnotesize{}89.81\%} & {\footnotesize{}77.54\%}\tabularnewline
\hline 
{\footnotesize{}\# somersaults} & {\footnotesize{}86.89\%} & {\footnotesize{}66.72\%}\tabularnewline
\hline 
{\footnotesize{}\# twists} & {\footnotesize{}95.15\%} & {\footnotesize{}93.51\%}\tabularnewline
\hline 
{\footnotesize{}Pose type} & {\footnotesize{}90.78\%} & {\footnotesize{}82.36\%}\tabularnewline
\hline 
\end{tabular}
\par\end{centering}
\caption{\label{tab:end-to-end-results}Combined classification accuracy.}
\end{table}

\FloatBarrier

\section{Conclusions}

There are challenges involved with composing multiple stages of deep
learning computer vision processing together. Using dive classification
as a case study, we have demonstrated that such a composite system
can be successfully constructed for sports action monitoring of continuous
video. Novel techniques for extracting action clips and localizing
an object of interest were presented with strong results. As future
work we would like to modify our system to assign dive scores like
a judge, which is a difficult problem due to the subtle and subjective
nature of the task.

\bibliographystyle{ieee}
\bibliography{Misc,Zotero}

\end{document}